\documentclass{Interspeech}

\interspeechcameraready

\title{Examining Test-Time Adaptation for Personalized Child Speech Recognition}

\author[affiliation={1,2}]{Zhonghao}{Shi}
\author[affiliation={1}]{Xuan}{Shi}
\author[affiliation={1}]{Anfeng}{Xu}
\author[affiliation={1}]{Tiantian}{Feng}
\author[affiliation={2,3}]{Harshvardhan}{Srivastava}
\author[affiliation={1}]{Shrikanth}{Narayanan}
\author[affiliation={1}]{Maja}{Mataric}

\affiliation{Viterbi School of Engineering}{University of Southern California}{United States}
\affiliation{}{Sara Technology Inc.}{United States}
\affiliation{Fu Foundation School of Engineering and Applied Science}{Columbia University}{United States}
\email{zhonghas@usc.edu,xuanshi@usc.edu, anfengxu@usc.edu, tiantiaf@usc.edu, hs3447@columbia.edu, shri@usc.edu, mataric@usc.edu}
\keywords{children speech recognition, human-computer interaction, test time adaptation}

\usepackage{comment}

\usepackage{multirow}
\usepackage{graphicx}
\usepackage{array}
\usepackage{booktabs}
\usepackage{bbm}
\usepackage{dsfont}
\usepackage[table]{xcolor}
\definecolor{lightblue}{HTML}{bfdcff}
\definecolor{lightgreen}{HTML}{bfdcff}
\definecolor{lightred}{HTML}{facece}

\usepackage{cite}
\usepackage{amsmath,amssymb,amsfonts}
\usepackage{algorithmic}
\usepackage{graphicx}
\usepackage{textcomp}
\usepackage{xcolor}
\usepackage{comment}
\def\BibTeX{{\rm B\kern-.05em{\sc i\kern-.025em b}\kern-.08em
    T\kern-.1667em\lower.7ex\hbox{E}\kern-.125emX}}

\begin{document}

\maketitle

\begin{abstract}

Automatic speech recognition (ASR) models often experience performance degradation due to data domain shifts introduced at test time, a challenge that is further amplified for child speakers. Test-time adaptation (TTA) methods have shown great potential in bridging this domain gap. However, the use of TTA to adapt ASR models to the individual differences in each child’s speech has not yet been systematically studied. In this work, we investigate the effectiveness of two widely used TTA methods--SUTA, SGEM--in adapting off-the-shelf ASR models and their fine-tuned versions for child speech recognition, with the goal of enabling continuous, unsupervised adaptation at test time. Our findings show that TTA significantly improves the performance of both off-the-shelf and fine-tuned ASR models, both on average and across individual child speakers, compared to unadapted baselines. However, while TTA helps adapt to individual variability, it may still be limited with non-linguistic child speech.

\end{abstract}

\section{Introduction}
\label{sec:intro}

Child-AI interaction enabled by AI software agents~\cite{huang2022chatbots} and socially assistive robots~\cite{belpaeme2013child} has shown great potential for many application domains, for example education~\cite{holmes2022state, li2023survey}. Conversational capabilities for these AI agents can support natural interaction with the child in achieving task goals \cite{Narayanan2002Creatingconversationalinterfacesfor}. To enable seamless human-like interaction, these AI agents and robots require accurate recognition of child speech \cite{Potamianos2003Robustrecognitionofchildrens}. 
Despite tremendous progress in machine learning methods for automatic speech recognition (ASR), a large body of recent work has shown that off-the-shelf ASR models do not generalize well to children's speech data, due to the high amount of acoustic and linguistic variability \cite{Lee1999Acousticsofchildrensspeech:}, resulting in data domain shifts 
between the adult data used for pre-training and 
for testing~\cite{bhardwaj2022automatic, GURUNATHSHIVAKUMAR2022101289}.

Recent work on child ASR~\cite{shraddha2022child, jain2023adaptation, thienpondt2022transfer, rolland2022multilingual, attia2024continued, rolland2024exploring, fan2022draft, liu2024sparsely, fan2024benchmarking} has experimented with various supervised and unsupervised methods to adapt off-the-shelf ASR models at training time. These studies have proposed applying methods such as transfer learning~\cite{shraddha2022child, jain2023adaptation, thienpondt2022transfer, rolland2022multilingual, attia2024kid}, continued pre-training~\cite{attia2024continued}, adapters~\cite{rolland2024exploring, fan2022draft}, low-rank adaptation~\cite{liu2024sparsely} and unsupervised domain adaptation~\cite{duan2021senone, duan2020unsupervised} to fine-tune and adapt the off-the-shelf ASR models with annotated children's speech data at training time. The results from these prior work have shown that both supervised and unsupervised adaptation at training time can substantially improve models for recognizing children's speech. 

\begin{table}[t!]
\caption{
Definitions of different model adaptation settings for child speech recognition, adapted from~\cite{wang2021tent}. Unlike other adaptation settings, TTA adapts off-the-shelf ASR models with $child^c$ data ($x^c$) only at test time,  without the need of annotations ($y^c$) or $adult^a$ pre-training dataset ($x^a$, $y^a$) at training time.
}
\vspace{-5mm}
\setlength{\tabcolsep}{1pt}
\label{tab:settings}
\begin{center}
\begin{tabular}{>{\centering\arraybackslash}m{3cm}>{\centering\arraybackslash}m{3cm}>{\centering\arraybackslash}m{1.5cm}}
\toprule
\bf Setting & \bf Train Loss & \bf Test Loss \\

\midrule
\shortstack{Supervised Pre-training}   & \multirow{1}{*}{$L(x^{a}, y^{a})$} & - \\
\midrule
\shortstack{Supervised Fine-Tuning}  &  $L(x^{c}, y^{c})$ & - \\
\midrule
\multirow{1}{*}{Unsupervised}  &  \multirow{2}{*}{$L(x^{a}, y^{a}, x^{c})$} & \multirow{2}{*}{-}  \\
\multirow{1}{*}{Domain Adaptation}  &  &  \\
\midrule

\multirow{1}{*}{Unsupervised}  &  \multirow{3}{*}{-} & \multirow{3}{*}{$L(x^{c})$}  \\
\multirow{1}{*}{Test-Time Adaptation}  & &  \\
\multirow{1}{*}{(Ours)}  & &  \\

\bottomrule
\end{tabular}
\vspace{-10mm}
\end{center}
\end{table}


Despite recent progress, training-time adaptation is not always feasible for real-world model deployments due to several limitations: (1) individual variability: each new child speaker introduces domain shifts at test time, and one-size-fits-all training-time adaptation fails to capture these individual differences; (2) annotation constraints: supervised adaptation requires labeled data, but obtaining annotations for each user poses significant financial, human labor, and logistical challenges; and (3) privacy and computational constraints: Users may prefer to keep their data private on local devices with limited computing and storage capacity, making both supervised and unsupervised training-time adaptation impractical.

Recent advancements in unsupervised test-time adaptation (TTA) methods offer a computationally efficient alternative~\cite{wang2021tent, lin2022listen, kim2023sgem}, enabling on-the-fly model adaptation to test data domains without requiring human annotations. However, no prior work has systematically evaluated the effectiveness of TTA methods in addressing domain adaptation challenges in child speech recognition. In this work, we investigate this gap by conducting experiments to answer three research questions, and report the following results and findings:



\begin{itemize}
\item \textit{RQ1: Why is TTA needed for ASR models in child speech recognition?} We found that both off-the-shelf and fine-tuned ASR models do not generalize robustly across different child speakers. Substantial domain shifts occur both across and within individual speakers, highlighting the need for adaptation at test time.

\item \textit{RQ2: Can unsupervised TTA methods effectively adapt both off-the-shelf and fine-tuned ASR models for child speech recognition?} As shown in Table~\ref{tab:tta-comparison}, we evaluated two popular TTA methods--SUTA and SGEM--across various off-the-shelf and fine-tuned model settings, using the MyST dataset. Our results indicate that SUTA is more robust than SGEM across different model settings. Notably, SUTA significantly improves both off-the-shelf models trained on out-of-domain adult speech and models fine-tuned on in-domain child speech. These findings further validate the need for test-time adaptation to account for individual variability.

\item \textit{RQ3: When do TTA methods perform well, and when do they fail?}
We conducted extensive analyses to probe the relationship between TTA performance and acoustic as well as linguistic characteristics of the input samples. We found that TTA may help adaptation to the individual background noise level, while non-linguistic speech may be particularly challenging and misleading for TTA's adaptation process.


\end{itemize}

\section{Methods}
\label{sec:methods}
In this work, we examine two widely used test-time adaptation (TTA) methods: 1) Single-Utterance Test-Time Adaptation (SUTA) ; and 2) Sequential-Level Generalized Entropy Minimization (SGEM). Our goal is to evaluate their effectiveness in adapting various off-the-shelf and fine-tuned ASR models to each child speaker in a continuous and unsupervised fashion.


\subsection{Problem Formulation}

A canonical ASR model can be denoted as $z = f(x, \theta)$, where $x$ is the input speech waveform, and $\theta$ refers to the ASR model parameters. 
Off-the-shelf ASR models are typically pre-trained on a primarily $adult^a$ dataset $D_{adult} = {\{(x_i^{a}, y_i^{a} )\}}_{I}$ in a supervised or self-supervised manner, to estimate $\theta^a$. The models are then used to transcribe $\hat{y}_j^{t}$ from $x_j^{t} \in D_{test}$, which assumes identical distribution with $D_{adult}$. However, when recognizing child speech $D_{child}$, there exists a significant data domain shift between $D_{adult} = {\{(x_i^{a}, y_i^{a} )\}}_{I}$ and $D_{child} = {\{(x_i^{c}, y_i^{c} )\}}_{I}$ due to the wide acoustic and linguistic variability in child speech and lack of child-specific ASR dataset  ~\cite{bhardwaj2022automatic,GURUNATHSHIVAKUMAR2022101289}.

To address this challenge, we use test-time adaptation (TTA) methods to adjust the parameters of the off-the-shelf ASR model $\theta^a \to \theta^c$, so that models can adapt to the domain shifts at test time for $child^c$ speech. The goal of TTA is to design optimization objectives ($\mathcal{L}$) based on the output logits $z \in \mathbb{R}^{L \times C}$ to adapt ASR models to the current test child's speech by continuously updating a small portion of model's parameters, where $z \in \mathbb{R}^{L \times C}$ is the predicted context logits. $L$ is the total number of timestamp, and $C$ is the number of word class.





\subsection{Test-Time Adaptation Methods}


This work experimented with two widely used TTA methods: 1) SUTA~\cite{lin2022listen}; and 2) SGEM~\cite{kim2023sgem} to adapt the feature extractor layers of Wav2vec2.0. The unsupervised optimization objective of SUTA~\cite{lin2022listen} consists of two parts: 1) Shannon entropy minimization loss ($\mathcal{L}_{em}$) and 2) negative sampling loss ($\mathcal{L}_{mcc}$). With the weighting hyper-parameter $\alpha$, the overall loss function is denoted as follows:

\begin{equation}
    \mathcal{L}_{SUTA} =  \alpha \mathcal{L}_{em} + (1-\alpha) \mathcal{L}_{mcc}
\end{equation}



On the other hand, the unsupervised optimization objective of SGEM~\cite{kim2023sgem} consists of two parts: 1) generalized R\'enyi entropy minimization ($\mathcal{L}_{\text{GEM}}$) and 2) negative sampling loss ($\mathcal{L}_{\text{NS}}$). With a weighting hyper-parameter $\lambda$, the overall loss function is denoted as follows:

\begin{equation}
\label{eq:overall}
\mathcal{L}_{SGEM}  = \mathcal{L}_{\text{GEM}} + \lambda \mathcal{L}_{\text{NS}},
\end{equation}

\subsection{Off-the-shelf and Fine-Tuned ASR Models}

We examined the effectiveness of TTA on two sets of models: 1) off-the-shelf models: off-the-shelf ASR models trained on LibriSpeech, which primarily include out-of-domain non-child speech data; 2) fine-tuned models: off-the-shelf ASR models fine-tuned on the training set of MyST with in-domain child speech data. Specifically, we used two off-the-shelf ASR models (Wav2vec2-base-960h\footnote{https://huggingface.co/facebook/wav2vec2-base-960h} and Wav2vec2-large-960h\footnote{https://huggingface.co/facebook/wav2vec2-large-960h}) that are trained on 960 hours of data from Librispeech~\cite{panayotov2015librispeech}. To further investigate the need for continuous adaptation to each child speaker’s characteristics after fine‐tuning, we fine‐tuned both Wav2vec2‐base and Wav2vec2‐large on the MyST dataset’s training set, which consists of child speech. We used a learning rate of $1e-5$ with the Adam optimizer for 50,000 steps, and a batch size of 32. The best model was selected based on the word error rate (WER) on the validation set, which was evaluated every 200 steps.

\subsection{Datasets}

This work used both the training and test set of My Science Tutor (MyST) dataset~\cite{ward2019my}, currently one of the largest publicly available datasets for child speech recognition\footnote{https://catalog.ldc.upenn.edu/LDC2021S05}. 
After removing recordings and utterances with missing annotations in the MyST dataset, we included data from 2622 children in the training set, and data from 438 children in the validation set, to fine-tune off-the-shelf ASR models. We used 91 childrens' data in the test set of MyST to evaluate TTA methods in all model settings, as listed in Table~\ref{tab:tta-comparison}. We only used utterances of less than 30s for the fine-tuning, following the setting in \cite{fan2024benchmarking}, resulting in 24.1 (SD=13.9) and 22.7 (SD=10.2) utterances for the training and validation dataset, respectively. For the test set, there were on average 140 (SD=99) utterances for each child speaker in the dataset. The duration of the utterances varied from less than 1 second to 111.4 seconds across all speakers. Overall, 129.3, 19.2, and 28.0 hours of data were included in the training, validation, and test set, respectively.

\subsection{Evaluation Experimental Setup}

We developed our codebase based on the open-sourced implementation from~\cite{kim2023sgem}~\footnote{https://github.com/drumpt/SGEM}. All experimental settings such as learning rate and optimizer were kept consistent between settings. For both TTA methods, the adaptation step was set to $N=10$. The weighting hyper-parameter was set to the following default values: $\alpha = 0.3$ for SUTA and $\lambda=0.3$ for SGEM. We evaluated ASR models using word error rate (WER). Due to the data distribution imbalance -- a small group of speakers had many utterances -- we report the unweighted average WER based on the speaker rather than the utterance, so the results accurately represent the ASR system's performance for each child.



\section{Experiments, Results, and Discussion}

\subsection{Why is TTA needed for ASR models in child speech recognition?}

\label{sec:rq1}
\begin{figure}[t!]
    \centering
    \includegraphics[width=1.0\linewidth]{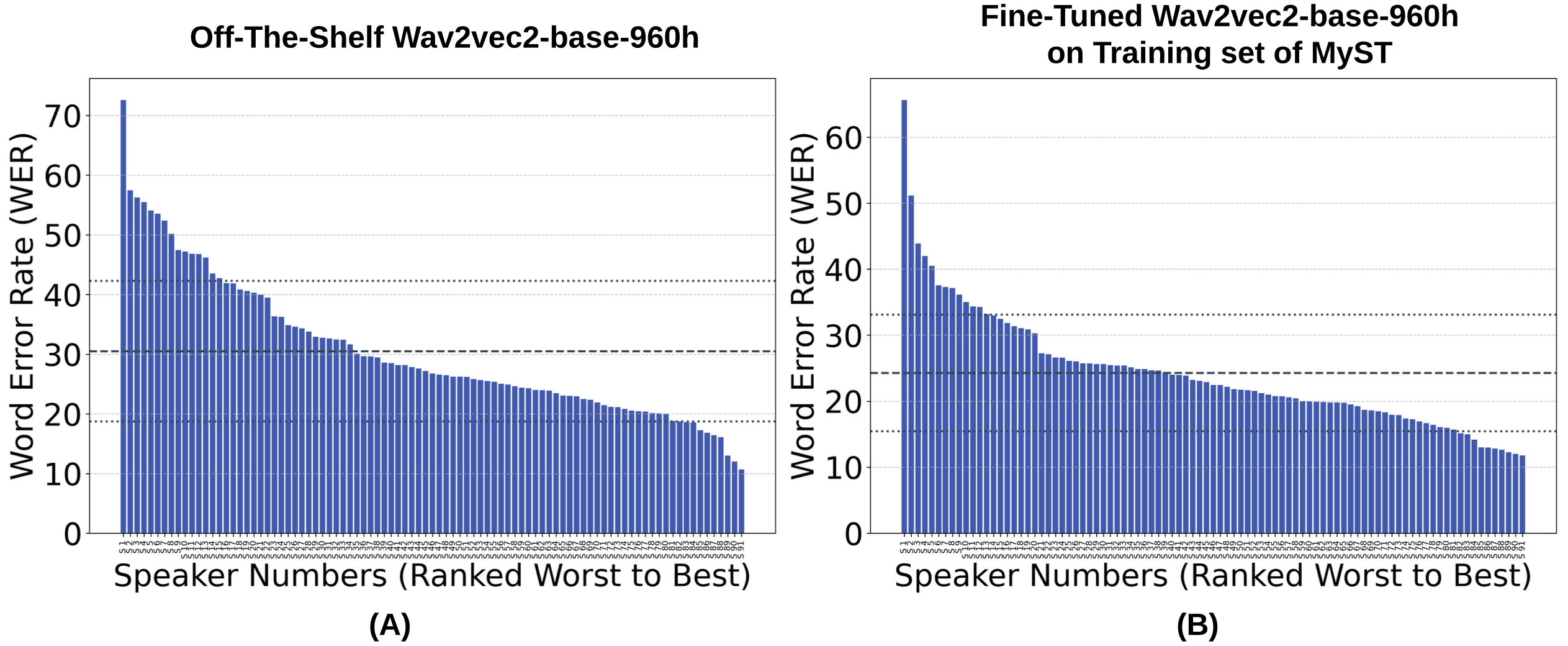}
    \vspace{-5mm}
    \caption{Unadapted baseline performance for both LibriSpeech-off-the-shelf and MyST-fine-tuned Wav2vec2-base models in word error rate (WER).}
    \label{fig:unadapted}
\end{figure}

Prior work has primarily reported average WER as a measure of model performance for child speech recognition~\cite{shraddha2022child, jain2023adaptation, thienpondt2022transfer, rolland2022multilingual, attia2024continued, rolland2024exploring, fan2022draft, liu2024sparsely, fan2024benchmarking}, but overall average WER tends to overlook individual WER differences between child speakers. As shown in Figure~\ref{fig:unadapted}, before we applied test-time adaptation (TTA) methods, we first analyzed the performance of off-the-shelf ASR models for each child speaker individually. As shown in Figure 1, for both unadapted off-the-shelf and fine-tuned Wav2vec2-base, we discovered substantial differences in WER across speakers. By fine-tuning the model with child speech (training set of MyST), we observed noticeable decreases in both WER average (from 30.5\% to 24.3\%) and standard deviation (from 11.8\% to 8.8\%) across individual speakers, indicating that fine-tuning is effective to improve ASR performance across child speakers. However, as visualized in Figure~\ref{fig:unadapted}(A) and (B), the differences quantified by the standard deviation still suggest that neither off-the-shelf nor fine-tuned ASR models generalize robustly across different child speakers, highlighting the need for more individual-level test-time adaptation to ensure model robustness for all speakers.

\begin{figure}[t!]
    \centering
    \includegraphics[width=\linewidth]{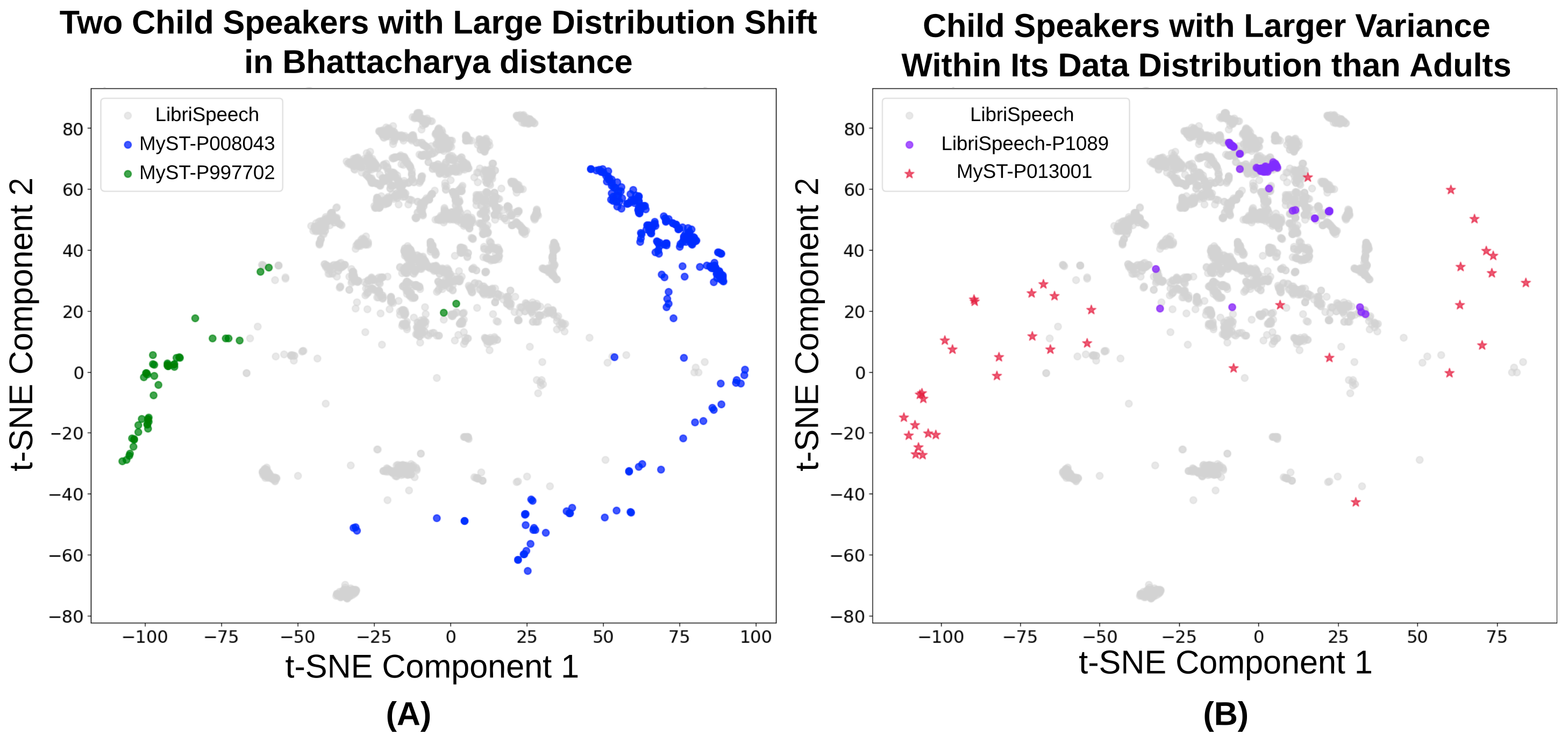}
    \caption{Domain shift across and within child speakers. (A) Significant domain shift between child speakers; (B) Significant variance and domain shift within child speakers.}
    \vspace{-5mm}
    \label{fig:tsne}
\end{figure}

We also analyzed and visualized the MFCC features extracted using Librosa~\cite{mcfee2015librosa} using the T-distributed stochastic neighbor embedding (t-SNE). We calculated the pair-wise Bhattacharya distance across all pairs of child speakers in MyST data, and visualized the pair with one of the largest distances in Figure~\ref{fig:tsne}(A). The results validate that there may be a significant distribution shift between child speakers. As shown in Figure~\ref{fig:tsne}(B), we also calculated the variance within each child speaker's embedding from MyST (Mean=1545.5 , SD=915.3) and found that it is significantly larger than the variance within each adult speaker from Librispeech (Mean=259.1, SD=275.0). Our findings suggest that child speakers may also have larger distribution shifts both across and within their data distribution due to more expressive and personal speech characteristics, unique non-linguistic and unintelligible speech, and environment noise, which need to be addressed at test time. These findings further motivate the need for applying test-time adaptation for child speech recognition to ensure robust model generalization to individual variability in real-world applications.

\begin{table}[t!]
\centering
\small 
\setlength{\tabcolsep}{4pt} 
\renewcommand{\arraystretch}{1.2} 
\caption{WER comparisons of TTA methods (SUTA and SGEM) with the unadapted baseline on test set of MyST dataset in different off-the-shelf and fine-tuned model settings.}
\begin{tabular}{lcccc}
\toprule
\textbf{Model Setting} & \textbf{TTA Method} & \textbf{WER} & \boldmath$\Delta$ & \textbf{Stat. Sig.}\\
\midrule
\multirow{3}{*}{\shortstack{\textit{Off-The-Shelf}\\Wav2vec2-base}} 
    & Unadapted          & 30.5\%           & --                & --            \\
    & SUTA               & 27.5\%           & -3\%              & \textit{p} \textless  .001 \\
    & \textbf{SGEM}      & \textbf{27.2\%}  & \textbf{-3.3\%}   & \textbf{\textit{p} \textless  .001} \\
\hline
\multirow{3}{*}{\shortstack{\textit{Off-The-Shelf}\\Wav2vec2-large}} 
    & Unadapted          & 26.6\%           & --                & --            \\
    & \textbf{SUTA}      & \textbf{25.1\%}  & \textbf{-1.5\%}    &  \textbf{\textit{p} \textless  .001}    \\
    & SGEM               & 25.3\%           & -1.3\%             &  \textit{p} \textless  .001  \\
\hline
\multirow{3}{*}{\shortstack{\textit{Fine-Tuned}\\Wav2vec2-base}} 
    & Unadapted          & 24.3\%           & --                & --            \\
    & \textbf{SUTA}      & \textbf{22.8\%}  & \textbf{-1.5\%}    & \textbf{\textit{p} \textless  .001}    \\
    & SGEM               & 23.8\%           & -0.5\%             & \textit{p} \textless  .001  \\
\hline
\multirow{3}{*}{\shortstack{\textit{Fine-Tuned}\\Wav2vec2-large}} 
    & Unadapted          & 23.2\%           & --                & --            \\
    & \textbf{SUTA}      & \textbf{22.5\%}  &  \textbf{-0.7\%}   & \textbf{\textit{p} \textless  .001}    \\
    & SGEM               & 23.4\%           & 0.2\%            & \textit{p} \textless  .001  \\
\bottomrule
\end{tabular}
\vspace{-7mm}
\label{tab:tta-comparison}
\end{table}

\subsection{Can unsupervised TTA methods effectively adapt both off-the-shelf and fine-tuned ASR models for child speech recognition?}
\label{sec:rq2}

\textbf{Off-the-shelf models.} As shown in Table~\ref{tab:tta-comparison}, on average, both SUTA and SGEM effectively reduced WER from the unadapted baseline. 
SUTA outperformed the unadapted baseline by 3\% in WER when using the off-the-shelf Wav2vec2-base, and SUTA outperformed the unadapted baseline by 1.5\% with the off-the-shelf Wav2vec2-large. In addition to reporting average model performance, we conducted individual two-sided Wilcoxon signed-rank tests between the TTA condition (N=91) and the unadapted condition (N=91), to further validate whether the improvements extend across all child speakers and are not driven by gains from a small group of speakers. We found that both SUTA and SGEM significantly improved from the unadapted baseline across different speakers for both Wav2vec2-base and Wav2vec2-large. 
These results suggest that TTA can help adapt ASR models from adult (LibriSpeech) to child (MyST) from the off-the-shelf models.

\begin{figure}[t!]
    \centering
    \includegraphics[width=1.0\linewidth]{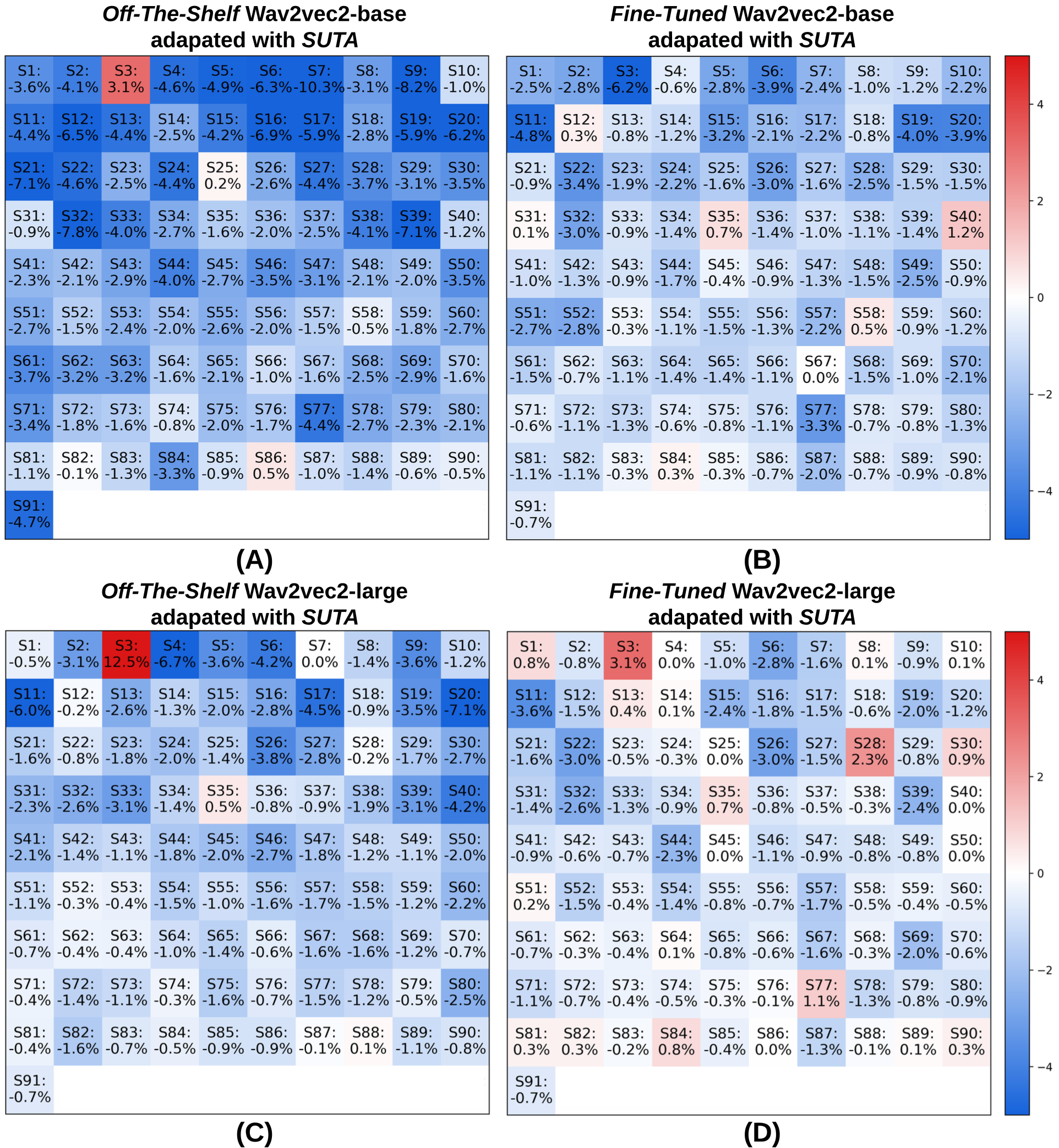}
    \caption{Heatplots of WER (\%) performance gains for all 91 child speakers in four model settings adapted by SUTA. Note: speakers are numbered (1-91) using the ranking listed in Figure~\ref{fig:unadapted} (A). }
    \label{fig:heatmap}
    \vspace{-3mm}
\end{figure}

On an individual speaker level, as shown in Figure~\ref{fig:heatmap} (A) and (C), we visualized the performance gain enabled by SUTA over the unadapted baseline for each child speaker. A darker blue indicates a larger performance gain from TTA, and the speakers are numbered using the ranking of unadapted WER presented in Figure~\ref{fig:unadapted} (A). For off-the-shelf models, the majority of speakers benefited from TTA. In Figure~\ref{fig:heatmap}(A), we found that S7, who benefited from SUTA the most when using off-the-shelf Wav2vec2-base, had a 10.3\% gain with SUTA. In Figure~\ref{fig:heatmap}(C), we found that S20, who benefited from SUTA the most when using off-the-shelf Wav2vec2-large, had a 7.1\% gain with SUTA. However, we also observed inconsistent cases such as S3 (WER drop of 3.1\%) in Figure~\ref{fig:heatmap}(A) and S3 (WER drop of 12.5\%), in Figure~\ref{fig:heatmap}(C). These results suggest that although TTA effectively improved WER for most speakers, the current TTA method may still not be sufficiently capable of providing performance gains for every child speaker.

\textbf{Fine-tuned models.} A body of prior work has shown that fine-tuning off-the-shelf models on in-domain child speech data can effectively adapt for the adult-child domain gap. However, based on our findings from Section~\ref{sec:rq1}, we hypothesized that the individual variability between each child speaker still needs to be adapted at test time. As shown in Table~\ref{tab:tta-comparison}, our results validated our hypothesis. Even after in-domain fine-tuning, on average, SUTA outperformed the unadapted baseline by 1.5\% in WER using the fine-tuned Wav2vec2-base, and by 0.7\% in the most challenging case of using the fine-tuned Wav2vec2-large. We noticed that SGEM performed worse than SUTA for the fine-tuned ASR models, suggesting that SUTA may be the more robust TTA method across different model settings. 

At the individual speaker level, as shown in Figure~\ref{fig:heatmap}(B) and (D), the majority of the speakers also benefited from TTA for fine-tuned models. However, the performance gains were less than the corresponding off-the-shelf models as expected. In Figure~\ref{fig:heatmap}(B), we found that S3, which benefited from SUTA the most when using the fine-tuned Wav2vec2-base, had a 6.2\% gain with SUTA after fine-tuning. In Figure~\ref{fig:heatmap}(D), we found that S11, which benefited from SUTA the most when using the off-the-shelf Wav2vec2-large, had a 3.6\% gain with SUTA after fine-tuning. However, we also observed performance drops, such as S40 (WER drop of 1.2\%) in Figure~\ref{fig:heatmap}(A) and S3 (WER drop of 3.1\%) in Figure~\ref{fig:heatmap}(D). When using the fine-tuned Wav2vec2-large, as shown in Figure~\ref{fig:heatmap}(D), 11 speakers experienced performance declines after adaptation with SUTA. While this result is expected given that fine-tuned Wav2vec2-large is the most accurate unadapted model setting, it further suggests that the current TTA methods may be less capable of selectively adapting to individual variability while preventing performance degradation for highly accurate models.

\begin{table}[t!]
\centering
\small 
\setlength{\tabcolsep}{4pt} 
\renewcommand{\arraystretch}{1.2} 
\caption{Acoustic and linguistic feature analysis. We conducted a Spearman’s rank‐order correlation test 
to examine the relationship between TTA’s performance gain and three acoustic and linguistic features: (1) effective mean squared (EMS) energy, and (2) word duration.}
\vspace{-1mm}
\begin{tabular}{lccc}
\toprule
\textbf{Model Setting} & \textbf{EMS Energy}  & \textbf{Word Duration}\\
\midrule
\shortstack{\textit{Off-the-shelf}\\Wav2vec2-base} & \shortstack{$\textbf{r = 0.22}$\\$\textbf{p = .04}$}               & \shortstack{$\textbf{r = -0.30}$\\$\textbf{p \textless\, .01}$}                       \\
\midrule

\shortstack{\textit{Off-the-shelf}\\Wav2vec2-large} & \shortstack{$r=0.02$\\$p > .05 $}              & \shortstack{$\textbf{r = -0.56}$\\$\textbf{p \textless\, .001}$}                       \\
\midrule

\shortstack{\textit{Fine-Tuned}\\Wav2vec2-base} & \shortstack{$r = -0.09$\\$p = 0.37 $}               & \shortstack{$\textbf{r = 0.41}$\\$\textbf{p \textless\, .001}$}                       \\
\midrule

\shortstack{\textit{Fine-Tuned}\\Wav2vec2-large} & \shortstack{$\textbf{r = 0.23}$\\$\textbf{p = 0.03} $}               & \shortstack{$\textbf{r = -0.29}$\\$\textbf{p \textless\, .01}$}                       \\
\bottomrule
\end{tabular}
\vspace{-5mm}
\label{tab:correlation}
\end{table}

\subsection{When do TTA methods perform well and when do they fail?}
\label{sec:rq3}


Background noise level can also be a source of individual variability that needs to be adapted at test time. To validate this, we conducted a Spearman’s rank-order correlation test with Holm-Bonferroni (HB) correction~\cite{holm1979simple} between speakers’ effective mean squared (EMS) energy level during non-speech regions and TTA’s performance gain. We obtained the non-speech regions using Silero voice activity detector~\cite{Silero_VAD} and calculated EMS energy using Librosa~\cite{mcfee2015librosa}. We found moderate correlations in both the off-the-shelf Wav2vec2-base and fine-tuned Wav2vec2-large model settings, suggesting TTA may help to adapt for background noise level at test time. 

In addition, we also examined the speech data of speakers who did not benefit from TTA, such as speaker S3 in Figure~\ref{fig:heatmap}(A), (C) and (D). Comparing them qualitatively with speakers who experienced greater improvements from TTA, we observed that those with performance degradation tended to produce a high proportion of prolonged and non-linguistic speech. Since these speech segments were not reflected in the ground-truth transcription, they may have misled the optimization process during TTA. Quantitatively analyzing these speech patterns would require human annotations to accurately identify non-linguistic speech. However, we noticed that such non-linguistic speech often occurred before or after linguistic speech, effectively increasing the overall duration of the audio clip beyond what was necessary. To approximate this effect, we calculated the duration of the word as:
\begin{equation*}
    \text{Word Duration} = \frac{\text{Total Speech Duration (seconds)}}{\text{Number of Words in Transcription}} 
\end{equation*}
Despite the limitations of this approximation, we found a significant correlation between speaking rate and TTA’s performance gain in all four modeling settings adapted by SUTA ($p \textless .01$) after HB correction, suggesting that the presence of non-linguistic speech may contribute to performance degradation in certain speakers.





\section{Conclusion}

In this work, we systematically examined the needs and performance of two widely used test-time adaptation (TTA) methods--SUTA, SGEM--to enable continuous adaptation of child speech data at test time. Our findings show that TTA significantly improved the performance of both off-the-shelf and their fine-tuned versions for child speech recognition, both on average and across individual child speakers, compared to unadapted baselines. Our acoustic and linguistic analysis also discovered that TTA may help adaptation to the individual background noise levels, while finding non-linguistic speech may be particularly challenging and misleading for test-time adaptation. 

\section{Acknowledgment}
This work was supported by National Science Foundation (IIS-1925083), Simons Foundation (SFI-AR-HUMAN-00004115-03, 655054), and Sara Technology Inc. The authors alone are responsible for the content and conclusions.



\bibliographystyle{IEEEtran}
\bibliography{mybib}

\end{document}